\documentclass[11pt]{article} %
\usepackage[preprint]{acl}

\usepackage{times}
\usepackage{latexsym}
\usepackage[T1]{fontenc}
\usepackage[utf8]{inputenc}
\usepackage{microtype}
\usepackage{inconsolata}
\usepackage{url}
\usepackage{booktabs}
\usepackage{graphicx}
\usepackage{multirow}
\usepackage{wrapfig}
\usepackage{pifont}
\usepackage{caption}
\usepackage{float}
\usepackage{afterpage}
\usepackage{subcaption}
\usepackage{xcolor}

\definecolor{darkblue}{rgb}{0, 0, 0.5}
\hypersetup{colorlinks=true, citecolor=darkblue, linkcolor=darkblue, urlcolor=darkblue}

\title{\projectname{}: An LLM-Driven Agent System for Autonomous \\Neuroscience Analysis}

\author{
  \textbf{Wuche Liu\textsuperscript{1}},
  \textbf{Yiran Qiao\textsuperscript{1}},
  \textbf{Linlin Hou\textsuperscript{1}},
  \textbf{Rui Yang\textsuperscript{1}},
  \textbf{Shusen Pu\textsuperscript{2}},
  \textbf{Song Wang\textsuperscript{3}},
  \textbf{Jing Ma\textsuperscript{1}}
  \\
  \textsuperscript{1}Department of Computer and Data Sciences, Case Western Reserve University, \\
  Cleveland, OH 44106, USA \\
  \textsuperscript{2}Department of Mathematics and Statistics, University of West Florida, \\
  Pensacola, FL 32514, USA \\
  \textsuperscript{3}Department of Computer Science, University of Central Florida, \\
  Orlando, FL 32816, USA
  \\
  \small{\textbf{Correspondence:} \href{mailto:wxl713@case.edu}{wxl713@case.edu},
  \href{mailto:jing.ma5@case.edu}{jing.ma5@case.edu}}
}

\newcommand{\projectname}{NS-Copilot}

\begin{document}

\maketitle

\begin{abstract}
AI is rapidly advancing neuroscience, yet many laboratories fail to fully unleash its potential
due to significant interdisciplinary barriers.
While pre-trained neural models for physiological data are progressing quickly, their
heterogeneous architectures and modality-specific constraints hinder systematic integration,
selection, and evaluation.
Despite recent advances in large language model (LLM)-based agent systems for intelligent
scientific applications, existing approaches often still lack the domain expertise required to
effectively select and coordinate diverse neuroscience pre-trained models and handle unique
data types in this domain.
We present \projectname{}, an LLM-driven multi-agent system for neuroscience analysis that
autonomously supports end-to-end workflows for diverse professional tasks.
It unifies domain-specific pre-trained models and supports key neuroscience modalities,
including EEG and extracellular spike data, through a natural-language interface.
Given raw data and a task description, \projectname{} orchestrates agents with specialized
roles for planning, adaptive control, code generation, and result synthesis, enabling analysis
without dataset-specific heuristics.
We evaluate \projectname{} on neuroscience benchmarks spanning Alzheimer's disease,
Parkinson's disease, and working memory spike decoding.
Across 8 trials per task, the system consistently outperforms strong baselines
on the primary metric, %
demonstrating the ability of \projectname{} for effective and scalable neuroscience analysis.
Our code is publicly available at \url{https://github.com/FrankLiu1102/ns-copilot}.
\end{abstract}

\section{Introduction}

Although AI is driving rapid progress in neuroscience, substantial interdisciplinary barriers
still limit the field's ability to fully realize its potential.
Pre-trained neural models for physiological data have advanced quickly
\citep{jiang2024largebrainmodellearning, ouahidi2025revefoundationmodeleeg, ye2025generalist, azabou2023unifiedscalableframeworkneural}, but their heterogeneous architectures and modality-specific constraints make systematic integration, model selection, and rigorous evaluation challenging \citep{xiong2025eegfmbench, kastrati2025eegbench}.
Meanwhile, despite recent progress in both scientific and general-purpose LLM-based agent
systems \citep{shen2023hugginggpt, huang2025biomni}, most primarily rely on code generation or
generic data analysis and machine learning toolkits which are often ill-suited for
domain-specific applications, and lack the domain expertise needed to effectively select and
coordinate diverse pre-trained neuroscience models and handle highly specialized data types.
This gap highlights the need for an
\textbf{autonomous system that can support end-to-end neuroscience analysis workflows.}

Despite its importance, this problem still presents three core challenges.
\textbf{First, bridging the interface gap between raw data and model-specific input requirements.} Many pre-trained neuroscience models expect data in a highly specific and domain-dependent format, which includes particular sampling rates, channel layouts, and tensor shapes.
This strict requirement reflects the deep architectural heterogeneity across the ecosystem.
Preparing raw recordings to satisfy these stringent constraints requires an extremely deep
understanding of both the underlying dataset structure and the target model's preprocessing
pipeline, posing a prohibitive barrier for non-expert users.
\textbf{Second, selecting the optimal method or model for a specific neuroscience dataset and task.} The ideal choice depends on a complex interplay of multiple factors: data modality (e.g., EEG vs.\ extracellular spikes), channel configuration, recording paradigm, and specific classification objectives.
Since no single pre-trained architecture dominates across all conditions, the
choice must be made per dataset. General-purpose AutoML systems \citep{erickson2020autogluon} do
not search over domain-specific pre-trained models, and EEG benchmarks
\citep{xiong2025eegfmbench} standardize evaluation but require manual per-model configuration.
\textbf{Third, adaptively optimizing the analytical workflow based on execution outcomes.} Even
when a suitable model is selected and the data is correctly formatted, the generated pipeline
may yield suboptimal performance due to misaligned hyperparameters, ineffective feature
engineering, or flawed evaluation strategies.
Recovering from such failures requires conducting a diagnosis of the root cause and determining
an optimization strategy to adaptively adjust the workflow, going beyond the code-level error
correction addressed by existing self-repair methods
\citep{chen2024selfdebugging, shinn2023reflexion}.

We propose \textbf{\projectname{}}, an autonomous LLM-driven agent system for neuroscience
analysis, which addresses the three aforementioned challenges through a unified pipeline,
requiring only raw neural data and a natural-language task description.
\projectname{} integrates multiple pre-trained neuroscience models in different applications.
To bridge the interface gap, a Coder agent automatically generates data preprocessing and model
invocation code at runtime based on declarative tool specifications (which describe the input
requirements of each model) inside our system.
To resolve model selection, a Planner agent automatically analyzes user input and dataset
metadata (modality, channel count, sampling rate, and class distribution), determines
appropriate evaluation metrics, and constructs a dynamically prioritized queue of compatible
models for selection.
To achieve adaptive optimization, a Controller agent analyzes execution metrics, diagnoses the
root causes of suboptimal results, and dynamically routes the system toward
targeted code patching (\textit{modify}) or full strategy resynthesis by the
Planner (\textit{replan}).
In the end, an Interpreter summarizes the final evaluation results.

In summary, our core contributions are as follows:
\begin{itemize}
\item \textbf{Unified pre-trained model orchestration framework.} We propose \projectname{}, a
novel LLM-driven agent system that first integrates nine representative pre-trained models
across different neuroscience data modalities and tasks, achieving fully automated neuroscience
analysis without dataset-specific heuristics.
\item \textbf{Automated model selection.} We design the Auto Model Selection protocol, in which
the Planner in \projectname{} automatically analyzes dataset characteristics and user intention
to determine the primary evaluation metric and construct a prioritized queue of compatible
models, with early stopping triggered once a candidate surpasses a performance threshold.
\item \textbf{Closed-loop diagnosis with explicit modify-vs-replan routing.} We design an
autonomous Controller that diagnoses suboptimal execution outcomes and
explicitly routes between targeted code patching (\textit{modify}) and full strategy
resynthesis (\textit{replan}), enabling iterative optimization without human intervention.
\item \textbf{Empirical validation across diverse neural paradigms.} We evaluate \projectname{}
on three benchmarks encompassing Alzheimer's disease, Parkinson's disease, and working memory
decoding, where it outperforms every baseline on the primary metric of
each.
\end{itemize}

\section{Related Work}

\paragraph{Neural foundation models for brain data.}
Pre-trained foundation models for neural recording data can be broadly categorized by their
target modalities.
For macroscopic EEG signals, a rapidly growing family of self-supervised models aims to learn
generalizable representations from large-scale clinical and research corpora.
Representative works include LaBraM \citep{jiang2024largebrainmodellearning}, CBraMod
\citep{wang2025cbramodcrisscrossbrainfoundation}, REVE
\citep{ouahidi2025revefoundationmodeleeg}, EEGPT \citep{wang2024eegpt}, EEGMamba
\citep{wang2025eegmamba}, and BrainOmni \citep{xiao2025brainomni}, with architectures ranging
from Transformers to state-space models, and pretraining scales from 2{,}500 hours to over
60{,}000 hours.
Models such as NeuroLM \citep{jiang2025neurolm}, BIOT \citep{yang2023biot}, and Neuro-GPT
\citep{cui2024neurogpt} further expand this ecosystem.
For extracellular spike recordings, models like NDT3 \citep{ye2025generalist}, MtM
\citep{zhang2024universaltranslatorneuraldynamics}, and POYO-1
\citep{azabou2023unifiedscalableframeworkneural} target multi-electrode array data, employing
autoregressive prediction, multi-task masking, and cross-attention mechanisms, respectively, to
learn transferable neural dynamics across sessions and brain regions.
Notably, NDT3 further demonstrates cross-species generalization spanning non-human primates and
human participants.
Despite this rapid progress, the ecosystem remains highly fragmented.
Each model possesses distinct input constraints, including specific channel montages, sampling
rates, and data formats, making cross-model comparisons laborious and error-prone.
Recent benchmarking efforts, such as EEG-FM-Bench \citep{xiong2025eegfmbench} and EEG-Bench
\citep{kastrati2025eegbench}, have begun standardizing the evaluation of EEG models; however,
these benchmarks require manual configuration on a per-model basis and do not cover the spike
modality.
Crucially, to our knowledge, no existing framework provides a unified interface to
automatically select, invoke, and evaluate these heterogeneous models based on the
characteristics of unseen datasets.

\paragraph{Scientific agents.}
Several systems have attempted to utilize LLMs as agents for scientific data analysis.
EEGAgent \citep{zhao2025eegagent} uses LLMs to orchestrate EEG analysis tools but does not
generate or execute code within a closed loop.
NeuroWeaver \citep{wang2026neuroweaver} formalizes EEG pipeline design as an evolutionary code
search problem with self-debugging, but it only targets a single modality and does not invoke
pre-trained foundation models.
SciDataCopilot \citep{rao2026scidatacopilot} provides an agentic data preparation framework for
EEG and MEG workflows, featuring code generation and execution capabilities, but its tool space
contains only preprocessing routines rather than foundation models.
Biomni \citep{huang2025biomni} is a general-purpose biomedical agent that integrates over 150
tools and supports retrieval-augmented planning and code execution, but it is oriented toward
broad biomedical tasks rather than neural recording analysis.
NOBEL \citep{tang2026nobel} unifies the representations of EEG, MEG, and fMRI within an LLM
backbone, but it exists as a single end-to-end model rather than an orchestrating agent for
multiple independent foundation models.
Concurrent systems SpikeAgent \citep{lin2025spikeagent} and BCI-Agent
\citep{marinllobet2025bciagent} apply LLM/VLM agents to spike sorting and cell-type classification, respectively, both with narrower scope. None of the aforementioned systems integrate multiple pre-trained neural foundation models in the neuroscience domain into a unified orchestration pipeline to achieve automated model selection, code generation, execution, and evaluation.

\paragraph{General agents.}
In the domain of general agents, AutoML systems such as AutoGluon
\citep{erickson2020autogluon}, FLAML \citep{wang2021flaml}, and Auto-sklearn
\citep{feurer2022autosklearn} automate model selection and hyperparameter tuning for
traditional machine learning models, but their search spaces do not include domain-specific
pre-trained foundation models.
HuggingGPT \citep{shen2023hugginggpt} and TaskMatrix.AI \citep{liang2023taskmatrix} demonstrate
LLM-driven selection of pre-trained expert models, yet their selection relies on task semantics
and model descriptions rather than the characteristics of the input data.
Regarding code self-repair, Self-Debugging \citep{chen2024selfdebugging} teaches LLMs to debug
their own code through execution feedback, Reflexion \citep{shinn2023reflexion} accumulates
experience across attempts using language reinforcement learning, and Self-Refine
\citep{madaan2023selfrefine} iteratively generates feedback and revisions without requiring
additional training.
These methods primarily address code-level errors (syntax and runtime exceptions).
Our Controller extends this paradigm to the analysis of execution metrics, diagnoses the root
causes of poor results, and explicitly chooses between targeted code patching and full
replanning.
Such modify-vs-replan routing has been explored in general ML-engineering agents
\citep{jiang2025aide,sun2023adaplanner}; to our knowledge, however, no LLM agent for closed-loop neuroscience data analysis adopts this routing as its Controller's core mechanism.

\section{Method}

\subsection{Overview}

\begin{figure*}[t]
\centering
\includegraphics[width=\textwidth]{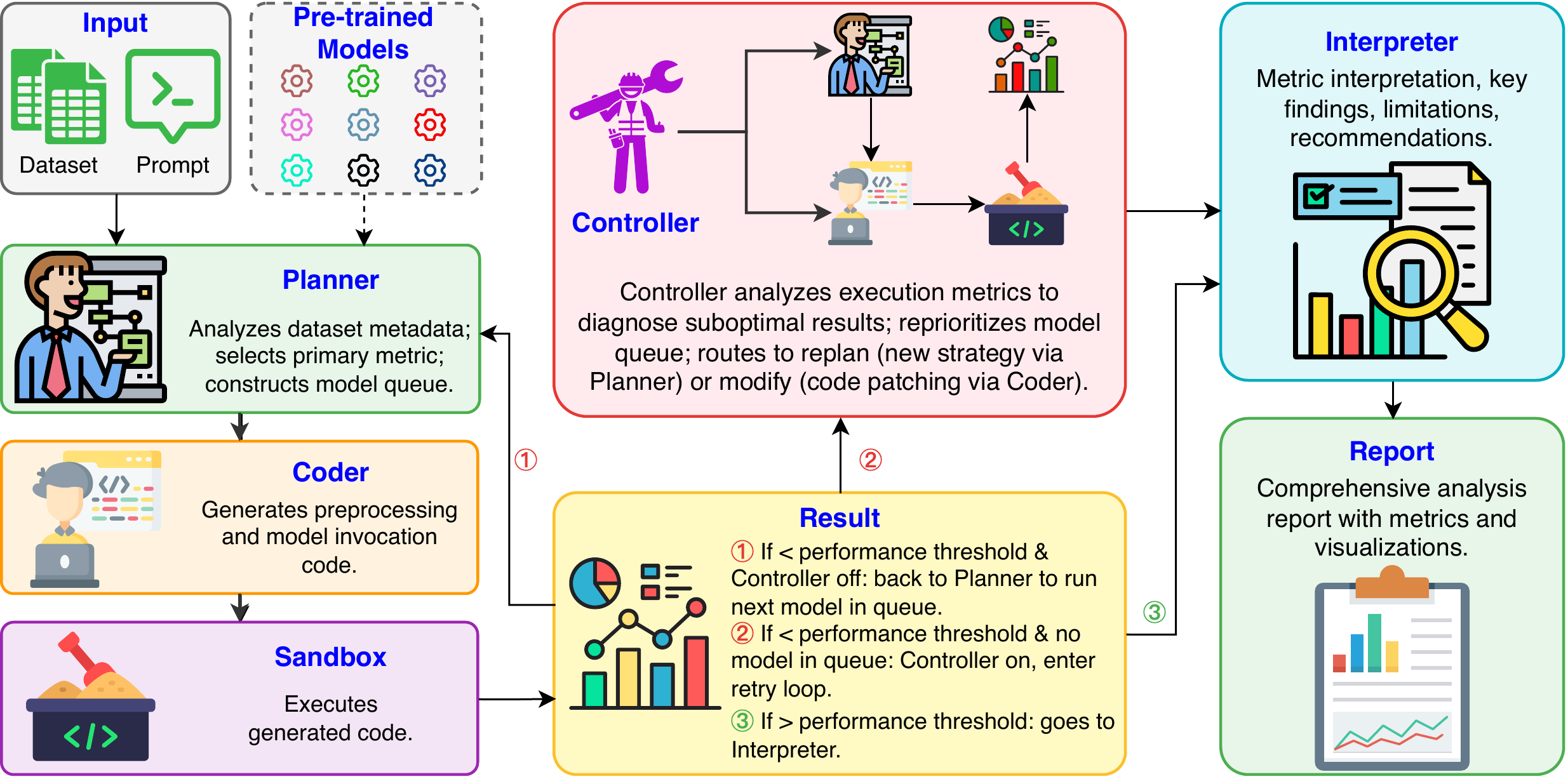}
\caption{Overview of the \projectname{} system. \textbf{Left column:} The user provides a dataset and a natural language prompt for task description. A registry of pre-trained models is available to the Planner via declarative tool specifications in the system. The Planner adaptively selects a primary metric, constructs a prioritized model queue, and generates an analysis plan. The Coder then synthesizes code for data preprocessing and model invocation, which is executed within a sandboxed environment. \textbf{Center column:} The \textit{Result} module compares execution outcomes against the performance threshold. If the current model fails to exceed this threshold and alternative models remain in the queue, the system proceeds to evaluate the next model (\ding{172}). If the entire queue is exhausted without surpassing the threshold, the Controller activates a closed-loop retry mechanism (\ding{173}). Here, it diagnoses suboptimal results, reprioritizes the model queue, and dynamically routes the execution to either a \textit{Replan phase} (synthesizing a new strategy via the Planner) or a \textit{Modify phase} (executing targeted code patches via the Coder). Conversely, if any result successfully exceeds the threshold, the pipeline immediately proceeds to the Interpreter (\ding{174}). \textbf{Right column:} The Interpreter produces a final comprehensive report detailing metric interpretations, key findings, limitations, and recommendations.}
\label{fig:pipeline}
\end{figure*}

\projectname{} is a fully automated multi-agent framework that enables users to complete
end-to-end neuroscience analyses by providing only raw neural data and a natural-language task
description.
As illustrated in Figure~\ref{fig:pipeline}, the system is collaboratively driven by four LLM
agents (\textit{Planner}, \textit{Coder}, \textit{Controller}, and \textit{Interpreter}).
Upon initiation, the \textit{Planner} analyzes the input to determine whether suitable
pre-trained models are available; if not, it generates code directly.
It then constructs a prioritized queue of pre-trained models based on dataset characteristics
and the user's intent.
The user may optionally specify a desired \textit{performance threshold}, which the system uses
as a stopping criterion to trigger early termination once the target is achieved.
If not provided, the system automatically determines an appropriate threshold.
Subsequently, the system sequentially runs each model within the prioritized queue, and early
stopping is triggered the moment the performance threshold is exceeded; however, if all models
fail to surpass the threshold, the \textit{Controller} intervenes to diagnose and optimize the
models sequentially.
Ultimately, the \textit{Interpreter} generates a comprehensive report containing metric
interpretations and recommendations.

\subsection{Planner}

The Planner serves as the orchestrating intelligence: it analyzes the structured dataset
profile (e.g., modality, channel count, sampling rate, and class distribution), dynamically
determines the optimal evaluation metric (e.g., utilizing balanced accuracy for imbalanced data
and macro F1 for multi-class classification tasks), and constructs a priority queue of
candidate models.
This ranking is derived from model metadata (such as pre-trained modality, channel
compatibility, and parameter scale) and is inferred entirely at runtime.

To satisfy users' performance requirements, the system incorporates a performance
threshold as a target criterion.
The core premise of the Auto Model Selection protocol in our system is that pre-trained models
do not always outperform carefully tuned traditional methods.
This performance threshold is also important to rigorously quantify the utility of pre-trained
models.
Users may specify this threshold directly; if none is provided, it is derived empirically: an
independent LLM programming agent (GPT-5.2-Codex; \citealp{openai2025codex}) receives the same
inputs to produce a competitive reference score.

Notably, Planner does not have access to the performance threshold score.
This architecture rules out metric selection bias from the Planner: the
choice of primary metric cannot be tuned to a score the Planner never observes, and is determined
from the dataset profile and the task description alone.
For each candidate model in the queue, the Planner formulates a targeted analysis plan (in JSON
format) based on dataset characteristics and model constraints, specifying the tools to be
used, the analytical steps, and the parameter configurations.

\subsection{Coder}

The Coder receives the analysis plan generated by the Planner, along with the dataset summary,
and automatically generates complete Python scripts based on the tool specification.
This process includes reading data from raw files, preprocessing to satisfy the input
requirements of the target model (e.g., resampling and channel selection), invoking the model's
inference interface to extract features, constructing a downstream classifier, and evaluating
performance.
The generated code is executed in a sandboxed environment, and all evaluation metrics are
automatically extracted.

\projectname{} currently integrates nine domain-specific pre-trained neural models across two
primary modalities, as shown in Table~\ref{tab:models}.
These models encompass a wide range of target tasks, architectures, embedding dimensions
(128 to 4096), and spatial constraints (ranging from a strict 10-20 montage to completely
channel-agnostic layouts).

\begin{table*}[t]
\centering
\caption{Integrated pre-trained models. \textit{Embed.}\ = dimension of the feature vector produced by each model's forward pass in our system (after any pooling or token concatenation; may differ from the model's internal hidden dimension); \textit{Ch.}\ = channel constraint; \textit{Hz} = target sampling rate; \textit{Std.\ pos.}\ = standard electrode positions; \textit{Downstream tasks} = representative evaluation tasks reported in the original paper. Models are grouped by modality: spike (top) and EEG (bottom).}\label{tab:models}
\resizebox{\textwidth}{!}{%
\begin{tabular}{llcccp{5.5cm}}
\toprule
\textbf{Model} & \textbf{Modality} & \textbf{Embed.} & \textbf{Ch.} & \textbf{Hz} & \textbf{Downstream Tasks} \\
\midrule
NDT3 \citep{ye2025generalist} & Spike & 1024 & --- & --- & Motor decoding (cursor control, reaching, bimanual) \\
MtM \citep{zhang2024universaltranslatorneuraldynamics} & Spike & 512 & --- & --- & Co-smoothing, forward prediction, choice decoding \\
POYO-1 \citep{azabou2023unifiedscalableframeworkneural} & Spike & 128 & --- & --- & Motor decoding (hand velocity, cross-session transfer) \\
\midrule
LaBraM \citep{jiang2024largebrainmodellearning} & EEG & 200 & 10-20 & 200 & Abnormal detection, event classification, emotion recognition \\
REVE \citep{ouahidi2025revefoundationmodeleeg} & EEG & 1536 & Any & 200 & Motor imagery, sleep staging, emotion recognition, abnormal detection \\
CBraMod \citep{wang2025cbramodcrisscrossbrainfoundation} & EEG & 600 & Any & 200 & Emotion recognition, motor imagery, sleep staging, seizure detection \\
EEGPT \citep{wang2024eegpt} & EEG & 2048 & 10-10 & 256 & Motor imagery, ERP detection, sleep staging \\
EEGMamba \citep{wang2025eegmamba} & EEG & 600 & Any & 200 & Emotion recognition, motor imagery, sleep staging, seizure detection \\
BrainOmni \citep{xiao2025brainomni} & EEG & 4096 & Std.\ pos. & 256 & AD/PD classification, emotion recognition, motor imagery, abnormal detection \\
\bottomrule
\end{tabular}%
}
\end{table*}

To manage this architectural diversity, all pre-trained models are abstracted behind a unified
tool API.
Each model is strictly exposed as a stateless forward-inference endpoint, responsible solely
for extracting latent representations; all subsequent downstream logic (such as temporal
pooling and classification) is dynamically synthesized by the Coder at runtime.
Each endpoint is accompanied by a structured specification (tool specification) detailing its
operational constraints, such as supported modalities, acceptable channel counts, and target
sampling rates.

\subsection{Controller}

If no candidate exceeds the performance threshold at the conclusion of the initial search
phase, the Controller agent initiates the closed-loop optimization phase.
Unlike standard LLM self-repair paradigms (which primarily focus on syntax errors or runtime
exceptions), our Controller performs deep semantic debugging.
By analyzing empirical execution metrics, it diagnoses the root causes of suboptimal
performance and dynamically routes between two retry strategies.

At the beginning of each retry round, the Controller dynamically rearranges the search queue,
elevating the priority of models that demonstrate the highest potential for improvement
relative to the threshold.
For each model in the updated queue, the Controller receives the complete execution trace: the
synthesized code, empirical metrics, the previous analysis plan, dataset metadata, and the
original user prompt.
Leveraging this rich context, it identifies subtle performance anomalies (such as accuracy
approaching random-chance levels or severe recall imbalances across classes) and formulates
highly specific, actionable recommendations.

The Controller translates the diagnostic findings into discrete routing decisions, selecting between two optimization pathways:

\textbf{Modify mode} is triggered when the high-level logic is sound but local algorithmic
adjustments are required (such as replacing the classifier, tuning regularization, or adjusting
the feature extraction window).
In this mode, the Planner is bypassed entirely; the previous scripts and the Controller's
targeted patching instructions are passed directly to the Coder.

\textbf{Replan mode} is invoked when empirical results indicate a fundamental flaw in the
methodological approach (such as a completely inappropriate feature extraction paradigm).
The diagnostic feedback is routed back to the Planner to synthesize an entirely new analysis
strategy from scratch.

Consistent with the initial search phase, if the optimized model exceeds the performance
threshold, early stopping is triggered immediately.
Otherwise, the Controller proceeds to optimize the next model in the queue until the threshold
is surpassed.
If all models in the queue fail to exceed the threshold after a full retry round, the system
enters the next retry round.
To balance computational overhead and optimization efficacy, the framework sets a default limit
of two retry rounds.

\subsection{Interpreter}

When the performance of a candidate model successfully exceeds the performance threshold
(whether during the initial search or the Controller retry phase), the Interpreter is invoked
to generate the final evaluation report.
It receives the complete execution results, including the evaluation metrics (accuracy,
balanced accuracy, macro F1, AUROC, and per-class recall), the confusion matrices, and the
generated visualizations, to scientifically interpret the outcomes, summarize key findings,
and propose recommendations for improvement.
If no model exceeds the threshold after all retry rounds, the framework reports the
best-performing model across all attempts along with its complete metrics.

\section{Experiments}

In this section, we present a comprehensive empirical evaluation of \projectname{} across three
distinct neuroscience benchmarks.
Specifically, our experiments are designed to answer the following core research questions. 
\begin{itemize}
    \item \textbf{RQ1}: How does \projectname{} perform compared with other agents and code generation baselines? 
    \item \textbf{RQ2}: To what extent does the Controller's closed-loop optimization contribute to the overall system performance? 
    \item \textbf{RQ3}: How do pre-trained models benefit the system? 
    \item \textbf{RQ4}: What are the primary failure modes of the whole system?
\end{itemize}

\subsection{Datasets}
\label{sec:datasets}

We evaluate \projectname{} on three publicly available neural datasets: (1)~\textbf{AD EEG}
(Alzheimer's Disease; OpenNeuro ds004504 \citep{ds004504}): resting-state EEG from 88 subjects,
19-channel 10-20 montage, 3-class classification (AD/FTD/Healthy); (2)~\textbf{PD EEG}
(Parkinson's Disease; OpenNeuro ds004584 \citep{ds004584}): resting-state EEG from 149
subjects, 64-channel 10-10 montage, binary classification (PD/Control); (3)~\textbf{WM Spikes}
(Working Memory; DANDI 000006 \citep{dandi000006}): extracellular spike recordings during a
delayed-response task, binary decoding of left vs.\ right lick decisions. We
use one session of 79 units and the 256 trials that remain after applying the dataset's own
quality flag and removing early-lick and no-response trials.
These benchmarks span three distinct neuroscience scenarios (Alzheimer's disease
classification, Parkinson's disease classification, and cognitive working memory decoding), two
recording modalities, diverse channel configurations (19--64 channels), and both binary and
multi-class tasks.

\paragraph{Data leakage audit.} Every pre-trained model is used as a frozen feature extractor, so if a benchmark
overlaps with a model's pre-training corpus, strong results may reflect prior exposure rather
than genuine transferability. Considering this, we audited all nine models against the three accession identifiers, and also at
the level of the underlying cohort, since one cohort can be released under several identifiers.
None of the three is part of any model's pre-training corpus; Appendix~\ref{app:leakage} gives
the per-model details. We note only one case of potential partial exposure.
BrainOmni uses ds004504 (same as ours in AD) as a downstream benchmark of its own, so its released
checkpoint was developed with visibility into results on that dataset. Its task there was a
two-class subset of 65 of the 88 subjects; however, ours is three classes over all 88, and the
backbone is frozen throughout our experiments: we do not fine-tune it on ds004504, so the exposure
is limited to what the released checkpoint already carries.

\subsection{Experimental Setup}
\label{sec:setup}

For each benchmark, we conduct 8 independent trials, reporting the mean $\pm$ standard
deviation across all metrics.
Each trial represents a fully autonomous, end-to-end execution, spanning from raw data
ingestion to final report generation without any manual intervention.

\paragraph{Evaluation metrics.} The Planner dynamically determines the primary optimization metric
based on dataset metadata: macro F1 for the 3-class AD task, and balanced accuracy for the
binary PD and WM tasks.
To provide a comprehensive evaluation, we additionally track accuracy, AUROC, and per-class
recall across all experiments.

\paragraph{Baselines.} We compare \projectname{} against three categories of baselines:
\textit{general agents} (AutoML-Agent \citep{automl_agent}, LAMBDA \citep{lambda_agent},
AutoGen \citep{autogen}), \textit{scientific agents} (SciAgents \citep{sciagents}), and
\textit{code generation baselines} (Claude-Sonnet-4.5 \citep{anthropic2025sonnet},
GPT-5.2-Codex \citep{openai2025codex}).
The Codex score establishes the performance threshold required to trigger early stopping.

\paragraph{Execution and computational budget.} All synthesized scripts are executed within a
sandboxed environment inside a Docker container, with restricted imports, controlled builtins,
and execution timeouts.
A single script execution is allowed 300\,s; a script that exceeds it is recorded as a
failed attempt and contributes a score of zero to that seed. This budget is uniform across models,
so it penalizes the models that need the most compute per trial, and Appendix~\ref{app:timeouts}
shows that one model, BrainOmni, is affected badly enough to distort its reported result.
If the initial search phase fails to yield a model that exceeds the baseline threshold, the
Controller's closed-loop optimization is strictly constrained to a default budget of two retry
iterations before termination.

\subsection{RQ1: Overall Performance}

Table~\ref{tab:main_results} summarizes the main results across all three benchmarks; the
complete per-model, per-stage results for AD and PD are given in Appendix~\ref{app:full-results}.
\projectname{} outperforms the state-of-the-art baselines on the primary metric of
all three datasets and on eight of the nine reported predictive metrics (the exception is AUROC on
WM), demonstrating that our agent system yields meaningful
gains through automatic planning and domain-specific pre-trained model
integration.\footnote{Below-chance baselines stem from method-specific failures (Appendix~\ref{app:baseline-failures}).} While \projectname{} exhibits a higher execution time due to multiple model trials and the overhead of running pre-trained models, this temporal cost is well justified by the resulting performance improvements and the complete elimination of manual, human-in-the-loop pipeline engineering. A complementary evaluation on a held-out test fold, run with queue search, early stopping and the Controller all switched off, is reported in Appendix~\ref{app:heldout}.

\begin{table*}[ht]
\centering
\caption{Performance comparison with baselines (mean $\pm$ std over 8 trials). \textbf{Primary Metric} auto-selected by the Planner: \textit{Macro F1} (AD), \textit{Balanced Accuracy} (PD/WM). \textit{Eff.}\ = avg execution time (s). Best in \textbf{bold}. \projectname{} uses \emph{online} early-stopping; per-seed winners: AD~=~REVE/BrainOmni/Vanilla, PD~=~Vanilla/LaBraM, WM~=~Vanilla/POYO-1.}\label{tab:main_results}
\resizebox{\textwidth}{!}{%
\begin{tabular}{ll cccc}
\toprule
\textbf{Dataset} & \textbf{Method} & \textbf{Primary Metric (\%)} & \textbf{Acc.\ (\%)} & \textbf{AUROC (\%)} & \textbf{Eff.\ (s)} \\
\midrule
\multirow{7}{*}{\shortstack[l]{AD EEG\\[-2pt]{\scriptsize (Macro F1$\uparrow$)}}}
  & AutoML-Agent \citep{automl_agent}  & $5.55 \pm 15.71$    & $6.95 \pm 19.64$    & $9.38 \pm 26.52$    & $255$   \\
  & LAMBDA \citep{lambda_agent}        & $43.57 \pm 38.28$   & $45.37 \pm 39.33$   & $54.99 \pm 45.83$   & $100$   \\
  & AutoGen \citep{autogen}            & $26.75 \pm 11.70$   & $32.75 \pm 10.59$   & $44.25 \pm 19.95$   & $103$   \\
  & Claude-Sonnet-4.5                  & $48.70 \pm 6.58$    & $51.00 \pm 7.18$    & $68.47 \pm 3.27$    & $304$   \\
  & Codex                              & $51.01 \pm 2.52$    & $53.98 \pm 2.98$    & $69.81 \pm 1.15$    & $68$    \\
  \cmidrule(l){2-6}
  & \projectname{}                     & $\mathbf{57.16 \pm 5.71}$ & $\mathbf{60.57 \pm 4.32}$ & $\mathbf{75.09 \pm 4.83}$ & $425$  \\
\midrule
\multirow{7}{*}{\shortstack[l]{PD EEG\\[-2pt]{\scriptsize (Bal.\ Acc.$\uparrow$)}}}
  & AutoML-Agent \citep{automl_agent}  & $17.39 \pm 32.29$   & $18.33 \pm 33.95$   & $19.90 \pm 37.01$   & $293$   \\
  & LAMBDA \citep{lambda_agent}        & $0.00 \pm 0.00$     & $0.00 \pm 0.00$     & $0.00 \pm 0.00$     & $284$   \\
  & AutoGen \citep{autogen}            & $6.25 \pm 17.68$    & $8.38 \pm 23.69$    & $6.25 \pm 17.68$    & $308$   \\
  & Claude-Sonnet-4.5                  & $65.12 \pm 3.27$    & $70.99 \pm 1.54$    & $74.07 \pm 3.52$    & $516$   \\
  & Codex                              & $65.52 \pm 7.98$    & $64.95 \pm 12.87$   & $71.07 \pm 8.23$    & $125$   \\
  \cmidrule(l){2-6}
  & \projectname{}                     & $\mathbf{71.32 \pm 2.97}$ & $\mathbf{71.91 \pm 1.87}$ & $\mathbf{78.15 \pm 1.21}$ & $975$ \\
\midrule
\multirow{8}{*}{\shortstack[l]{WM Spikes\\[-2pt]{\scriptsize (Bal.\ Acc.$\uparrow$)}}}
  & AutoML-Agent \citep{automl_agent}  & $18.88 \pm 26.05$   & $20.62 \pm 28.48$   & $19.00 \pm 26.23$    & $196$   \\
  & LAMBDA \citep{lambda_agent}        & $70.62 \pm 3.11$    & $72.00 \pm 3.66$    & $78.50 \pm 4.38$    & $66$    \\
  & AutoGen \citep{autogen}            & $55.00 \pm 0.00$    & $56.00 \pm 0.00$    & $59.00 \pm 0.00$    & $20$    \\
  & SciAgents \citep{sciagents}        & $50.00 \pm 0.00$    & $51.00 \pm 0.00$    & $59.00 \pm 0.00$    & $40$    \\
  & Claude-Sonnet-4.5                  & $73.97 \pm 6.99$    & $74.76 \pm 7.05$    & $\mathbf{83.89 \pm 6.70}$    & $164$   \\
  & Codex                              & $72.58 \pm 1.68$    & $72.77 \pm 1.94$    & $78.87 \pm 2.23$    & $97$    \\
  \cmidrule(l){2-6}
  & \projectname{}                     & $\mathbf{74.43 \pm 3.18}$ & $\mathbf{74.81 \pm 2.93}$ & $80.93 \pm 4.22$ & $470$  \\
\bottomrule
\end{tabular}%
}
\end{table*}

\subsection{RQ2: Effect of the Controller}

To evaluate the contribution of the Controller module, we measure the performance of our system
across three distinct stages: (1)~\textbf{Initial Run}, where the Planner generates an analysis
plan and the Coder produces executable code with up to three self-correction attempts for
runtime errors, yet without Controller intervention; (2)~\textbf{+\,Retry\,1}, where the
Controller diagnoses the Initial result by identifying anomalies and issuing targeted
recommendations, subsequently triggering a full replan-or-modify cycle; and
(3)~\textbf{+\,Retry\,2}, where the Controller diagnoses the execution once more to trigger a
second optimization cycle.
We run this ablation on AD, over Vanilla and five pre-trained models (BrainOmni is
excluded; see the caption), with the same 8 independent seeds for each configuration (48 runs).
Table~\ref{tab:ablation_controller} reports the change in Macro F1 relative to the Initial Run for
the forced Retry~2 round and for the value the system returns, together with the seed-to-seed
standard deviation of Retry~2 and of the returned value.

\vspace{3mm}
\begin{table*}[!t]
\centering
\small
\caption{Ablation of the Controller on AD, over all six pre-trained and vanilla
configurations and eight seeds each (48 runs). $\Delta$ \textit{(exploration)} is Retry~2 minus the
initial round: what two forced rounds change on their own. $\Delta$ \textit{(returned)} is what the
system reports, the best of the three rounds for each seed, minus the initial round; it cannot be
negative by construction, so the quantity of interest is its size, not its sign. The last two
columns give the seed-to-seed standard deviation of Retry~2 and of the returned value.
\textbf{Bold} marks the three configurations where selecting the best round left the
output less dispersed than Retry~2.
BrainOmni is excluded: its AD figures are distorted by execution timeouts
(Appendix~\ref{app:timeouts}).}
\label{tab:ablation_controller}
\begin{tabular}{lcccc}
\toprule
\textbf{Configuration} & \textbf{$\Delta$ (exploration)} & \textbf{$\Delta$ (returned)} & \textbf{Retry~2 (std)} & \textbf{Returned (std)} \\
\midrule
Vanilla   & $-2.32$ & $+2.54$ & $9.53$  & $\mathbf{4.97}$ \\
LaBraM    & $-2.18$ & $+3.50$ & $3.79$  & $6.71$ \\
REVE      & $-2.91$ & $+0.65$ & $5.24$  & $\mathbf{3.40}$ \\
CBraMod   & $+3.48$ & $+7.38$ & $6.91$  & $9.12$ \\
EEGPT     & $-2.36$ & $+1.22$ & $11.00$ & $\mathbf{3.65}$ \\
EEGMamba  & $+5.62$ & $+7.51$ & $6.53$  & $6.79$ \\
\midrule
Mean & $-0.11$ & $+3.80$ & & \\
Runs improved / degraded (rest unchanged) & $20$ / $20$ & $25$ / $0$ & & \\
\bottomrule
\end{tabular}
\end{table*}

Table~\ref{tab:ablation_controller} separates two things the Controller does. The
forced exploration is close to neutral on AD: two retry rounds move the mean by $-0.11$\,pp and
improve as many runs as they degrade ($20$ against $20$ of $48$). What the system returns, the best
of the three rounds for each seed, is $+3.80$\,pp above the initial round and is worse than it on
none of the $48$ runs. A retry round is a fresh point in the configuration
space rather than a refinement of the previous one: Modify mode substitutes the classifier, the
regularization or the feature window, and Replan mode discards the previous plan and synthesizes a
new strategy from scratch. The Controller routes between the two on the evidence of a single score,
and when the shortfall is small that score carries more noise than signal; nothing requires the new
result to beat the old one before it is recorded. A round that lands on a worse point is therefore 
an expected cost of exploring. The Controller's contribution lies in retaining the best round rather than in the exploration itself, and the two $\Delta$ columns are designed to disentangle these effects.
This also explains the standard deviations. They grow across the forced rounds because
exploration is genuinely erratic: for each of the six configurations, the largest per-stage
standard deviation of Macro F1 in Table~\ref{tab:ad_full} falls in Retry~1 or Retry~2 rather than
in the initial round. One EEGPT seed collapses in Retry~2, which alone
lifts that round's standard deviation to $11.00$. The selection rule discards that round, and the
returned standard deviation is $3.65$. The dispersion
of the forced rounds describes the search, not the output. The forced exploration is more clearly harmful on PD, where
Table~\ref{tab:pd_full} puts Retry~2 an average of $-2.34$\,pp in Balanced Accuracy below the
initial round over the same six configurations.

\subsection{RQ3: Effect of Pre-trained Models}

To quantify the benefit of pre-trained neural models, we compare the vanilla pipeline
(handcrafted spike-count features) against POYO-1 on the WM
task, with the model fixed in advance and both arms scored on the
same held-out partition for each of the 8 seeds.
POYO-1 yields a consistent gain of $+6.82$\,pp in Balanced Accuracy on the
held-out evaluation ($75.30 \pm 8.39$ vs.\ $68.48 \pm 7.49$), favoring the pre-trained model on
7 of 8 seeds (Table~\ref{tab:heldout}).
This improvement is consistent with POYO-1's PerceiverIO architecture, which is
designed for variable neuron counts; our comparison is end-to-end and does not isolate that
property.
These results indicate that pre-trained models can provide meaningful
representational advantages over classical features, and that the size of the advantage varies by
task, from $+13.34$\,pp on AD to a near-tie on PD ($+0.68$; Table~\ref{tab:heldout}).
It is that variation which motivates the automated model selection protocol at the core of
\projectname{}.
It is worth being explicit about where the advantage comes from. Codex and
Claude-Sonnet-4.5 in Table~\ref{tab:main_results} are pure LLM agents: they receive the same task
description and write their own analysis code, with no pre-trained neural model available. The
\textit{Vanilla} configuration is our architecture with those models removed, and on
the two benchmarks with full per-stage results its initial round scores \emph{below} Codex ($49.80$
vs.\ $51.01$ on AD and $64.52$ vs.\ $65.52$ on PD; Tables~\ref{tab:ad_full}
and~\ref{tab:pd_full}). Orchestration on its own therefore does not beat a single-shot
LLM agent. What the comparison isolates is the integration itself: making nine pre-trained models
addressable through one interface, and choosing among them per task, is what produces the gain.
The held-out evaluation of Appendix~\ref{app:heldout} is also
what separates that gain from adaptive selection: there the queue search, the early stopping and
the Controller retry loop are all switched off and the model is fixed in advance, so no part of the reported advantage can come from
choosing the winner after the fact.

\subsection{RQ4: Error Analysis}

\begin{figure}[t]
\centering
\includegraphics[width=0.8\columnwidth]{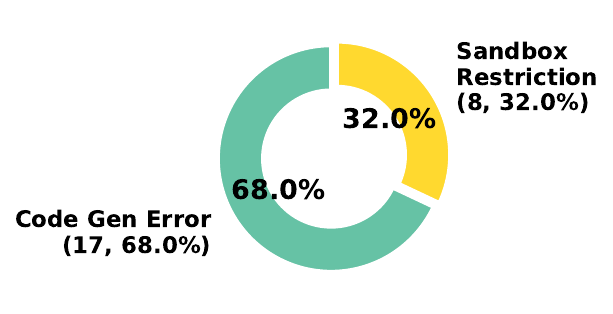}
\caption{Error distribution: 25 code-level failures out of 325 attempts (92.3\% code-level success rate).}\label{fig:errors}
\end{figure}

To systematically characterize the failure modes of our system, we analyze 325
execution attempts across the three benchmarks; a script that exceeds the 300\,s budget is
recorded as a failed attempt in the reported scores and is examined in
Appendix~\ref{app:timeouts} rather than counted here.
The system achieves a 92.3\% code-level success rate, leaving 25 code-level failures distributed across
two primary categories (Figure~\ref{fig:errors}).
Code generation errors (68.0\%) arise when the LLM produces scripts containing incorrect
variable references or label format mismatches.
Sandbox module restrictions (32.0\%) occur when the generated code attempts to import libraries
not permitted within the isolated execution environment.
Both failure modes are non-catastrophic: they surface as explicit execution
errors rather than as silently incorrect results, and a failure costs a candidate slot rather than
the run, since the system continues through the prioritized queue and reports the best model
across all attempts.

\section{Conclusion}

We propose \projectname{}, an autonomous LLM-driven agent system that unifies nine pre-trained
models across EEG and spike modalities behind a single natural-language interface, integrating
automated model selection, code generation, and closed-loop optimization.
Across three benchmarks, \projectname{} outperforms every baseline on the
primary metric; ablations show that the gain comes from the pre-trained models rather than from
orchestration alone, that the Controller's value lies in keeping the best round rather than in
the forced exploration, and that the optimal model varies across tasks.
By lowering the barrier to advanced neuroscience analysis, \projectname{} enables researchers
without extensive AI expertise to leverage state-of-the-art models and accelerate data-driven
discovery.

\clearpage

\section*{Limitations}

Current limitations include increased execution time and computational cost due to
closed-loop evaluation, which is necessary to reliably explore candidate models and select the
optimal method for each task.
Instrumented per-agent token measurements indicate that \projectname{}'s per-run cost is
roughly 1.5--2$\times$ that of the single-step Codex baseline; this overhead is
justified by the performance gains over Codex. Each of the three scenarios is represented by a single dataset. Adding further datasets within each scenario, so that the same protocol meets greater within-modality variation, is the direction in which we intend to grow the benchmark. Future work will extend the framework to incorporate additional neuroscience tools, add a dedicated validator agent that checks label alignment and output shapes, support new modalities such as fMRI and MEG, explore efficient orchestration strategies, and enable more complex, multi-step analysis workflows.

\section*{Ethics Statement}

\paragraph{Use of Large Language Models.}
This work uses LLMs as a core component of the proposed system: an LLM serves as the
orchestration agent that selects pre-trained models, generates analysis code, and drives
iterative optimization.
Additionally, Claude (Anthropic) was used to assist with code development and paper
grammar-checking during this research.
LLMs were not used to generate any original research ideas or any original content.
All LLM-generated content was reviewed and verified by the authors.

\paragraph{Datasets.}
All datasets used in this work are publicly available under open licenses.
The Alzheimer's/FTD EEG dataset (OpenNeuro ds004504) is released under CC0 (Public Domain).
The Parkinson's disease EEG dataset (OpenNeuro ds004584) is released under CC0 (Public Domain).
The working memory spike dataset (DANDI 000006) is released under CC-BY 4.0.
No private or restricted-access data was used.
All datasets were collected by their original authors with appropriate ethical approvals as
described in their respective publications.

\paragraph{Pre-trained Models.}
The nine integrated pre-trained models are publicly available for research use.
All model code is released under permissive open-source licenses (MIT or Apache 2.0).
Model weights are distributed by their respective authors via GitHub, Hugging Face, or
dedicated servers under varying terms: some under permissive licenses (MIT or Apache 2.0), some
under non-commercial licenses (CC-BY-NC 4.0), and some with custom access agreements.
REVE weights require accepting a gated access agreement on Hugging Face.
Our framework does not redistribute any model weights; users download them directly from the
original sources.
We obtained access to all models and used them solely for non-commercial academic research in
compliance with their respective terms.

Overall, this work does not raise any significant ethical concerns.

\paragraph{Broader Impact.}
Our framework is designed to lower the barrier for neuroscientists to leverage AI agents and
domain-specific pre-trained models.
While this democratization of access is broadly positive, we note that automated analysis
pipelines should complement, not replace, domain expertise.
Users should critically evaluate all outputs, particularly in clinical contexts where incorrect
classifications could have downstream consequences.
Overall, \projectname{} represents a step toward more accessible, scalable, and trustworthy
AI-driven neuroscience research.

\section*{Reproducibility Statement}

All source code, including the complete multi-agent pipeline, tool specifications, and
experiment scripts, is publicly available at
\url{https://github.com/FrankLiu1102/ns-copilot}.
To ensure reproducibility, the system is fully containerized via Docker (including the
Dockerfile and Docker Compose configuration), enabling a seamless, single-command environment
setup; Appendix~\ref{app:web-ui} shows the web interface it exposes.
Experimental reproduction is facilitated through \texttt{run\_test.sh}, which
fixes the random seeds each run uses. Because the analysis code is written by the LLM agents at run
time, a repeated run is not guaranteed to reproduce a given score exactly; every result we report
is therefore a mean and standard deviation over eight seeds rather than a single figure.
All datasets are publicly available: the AD EEG dataset (OpenNeuro ds004504, CC0), the PD EEG
dataset (OpenNeuro ds004584, CC0), and the WM spike dataset (DANDI 000006, CC-BY 4.0).
Furthermore, the weights for all nine pre-trained models are publicly accessible, and
\texttt{setup.sh} downloads eight of them automatically once the user has accepted the access terms
set by the individual providers; the EEGPT checkpoint must be downloaded manually from Figshare,
which blocks automated retrieval.
The specific LLMs utilized in this study are GPT-5.2 (OpenAI, powering all four NS-Copilot
agents), GPT-5.2-Codex (OpenAI, employed to establish the performance threshold), and Claude
Sonnet 4.5 (Anthropic, serving as one of the baselines).
Each agent baseline (AutoML-Agent, LAMBDA, AutoGen, SciAgents) was evaluated using its
originally published default LLM configuration.

\bibliography{emnlp2026}
\bibliographystyle{acl_natbib}

\appendix

\section{Baseline Failure Modes}
\label{app:baseline-failures}

Several baselines exhibit below-chance performance, or produce no
usable result at all, on AD, PD and WM; we attribute these to
method-specific failures rather than unfair comparison.
AutoML-Agent's rigid Optuna-based search proved incompatible with our raw EEG inputs, causing pipeline crashes on 7/8 AD trials (returning 0). On the WM spike benchmark AutoML-Agent likewise failed to produce runnable code on 5 of 8 trials, which are scored as 0 and pull the mean below chance.
AutoGen consistently generated severely biased pipelines (e.g., 52\% AD recall but 5\% FTD recall on the 3-class task, indicating majority-class collapse).
SciAgents, designed for hypothesis generation rather than supervised classification,
produced a degenerate chance-level result on WM, the only benchmark on
which we report it: its balanced accuracy is 50.00\%, the chance level for a binary task, and its
standard deviation is zero on all three reported metrics across the eight trials
(Table~\ref{tab:main_results}).
LAMBDA did not produce a parseable result on any of the PD trials we report:
its sessions ended inside its own self-repair loop without printing final metrics, and such a
trial contributes a score of zero under the same convention we apply to our own runs
(Section~\ref{sec:setup}).
Importantly, these failures are themselves informative: they demonstrate that domain-aware orchestration, rather than a stronger generic LLM agent, is what enables consistent autonomous performance in this domain.

\section{Data Leakage Audit}
\label{app:leakage}

None of our three benchmarks (ds004504, ds004584, DANDI 000006) is part of the
pre-training corpus of any of the nine integrated models. We matched on accession identifiers
rather than dataset names, which are not unique.

Identifiers alone are not a sufficient audit, because a single cohort can be released under
several of them, and the corpora themselves demonstrate this: BrainOmni pre-trains on the
Healthy Brain Network cohort under both ds004186 and ds005505 to ds005512. We therefore also
checked the alternative releases of our own cohorts, namely ds006036 for the AD cohort and
ds004574, ds004579 and ds004580 from the laboratory that released ds004584. None appears in any
corpus.

Two of the nine models draw on OpenNeuro, and both enumerate what they use: REVE lists 55
accession identifiers and BrainOmni 24, and neither list contains ours. The nearest is ds004582,
which differs from ds004584 only in its final digit. Clinical cohorts of the same two diseases do
occur in these corpora and should not be mistaken for ours: REVE and BrainOmni both pre-train on
ds002778, a separate Parkinson's resting-state set from UC San Diego, and BrainOmni on ds004796,
an Alzheimer's-risk cohort.

The remaining EEG models pre-train on TUEG, PhysioNet, SEED and comparable corpora, which do not
redistribute OpenNeuro recordings. Among the spike models, POYO-1 and NDT3 pre-train on
non-human primate motor cortex; the NDT3 checkpoint we load is the 200-hour configuration, whose
released dataset list contains eleven monkey reaching corpora and no rodent or human source. MtM
pre-trains on mouse recordings from other brain regions collected by a different laboratory.

The one partial exposure noted in Section~\ref{sec:datasets} is BrainOmni's use of ds004504 as
its own downstream benchmark, where it appears as a 65-subject two-class subset of the 88
subjects we use. Its authors state that all datasets used for downstream testing were excluded
from pre-training, so the pre-training claim above holds, but the checkpoint's design choices
were made with those results visible, and we therefore do not describe it as fully naive to that
benchmark.

\section{Held-out Evaluation}
\label{app:heldout}

A cross-validation score that is also used to steer model selection and retries
becomes optimistically biased. To separate the contribution of the pre-trained models from that
selection process, we re-ran each task with the selection process removed entirely and with the
final score taken on data that no part of the run had touched. Table~\ref{tab:heldout} reports the
outcome.

Two points bound what this evaluation shows. It compares one pre-trained model against
\textit{Vanilla}, the identical pipeline with the pre-trained models withheld, so the only variable
is whether a pre-trained model was used; it is not a comparison against the single-step LLM
baselines. And it does not test whether the queue would have selected the strongest model, since
the model is fixed in advance rather than chosen.

The PD row is worth reading carefully. An earlier run of this experiment on three seeds put the PD
gap at $+5.14$; all three happened to favor BrainOmni. At eight seeds the gap is $+0.68$ and five
of eight seeds favor it, which is the near-tie reported here. This does not
contradict the PD result in Table~\ref{tab:main_results}, a clear gain over Codex: that
table reports the full system with selection enabled, which is a different
comparison from the one made here, where a single model is held fixed against \textit{Vanilla} on
a fold no part of the run has touched. The drop from three seeds to eight is a concrete instance
of the small-sample effect that motivated this evaluation in the first place.

\begin{table*}[!t]
\centering
\small
\caption{Held-out evaluation, $n=8$ seeds (5678, 6543, 7890, 42, 123, 456, 2048,
4096). Stratified 70/15/15 train/validation/test split; every hyperparameter is selected by
cross-validation on the training portion alone, and the test fold is read once, at the end. Queue
search, early stopping and the Controller retry loop are switched off for these runs, which
consist of a single initial round with the model fixed in advance, so no advantage reported here
can arise from adaptive selection. The comparison is paired: for a given seed both arms use the
same partition. \textit{Seeds} counts how many of the eight favor the pre-trained model.}\label{tab:heldout}
\resizebox{\textwidth}{!}{%
\begin{tabular}{ll cc c c}
\toprule
\textbf{Dataset} & \textbf{Metric} & \textbf{Pre-trained model} & \textbf{Vanilla} & \textbf{$\Delta$} & \textbf{Seeds} \\
\midrule
AD & Macro F1 & $57.71 \pm 16.22$ (REVE) & $44.37 \pm 11.01$ & $+13.34$ & 6/8 \\
WM & Bal.\ Acc. & $75.30 \pm 8.39$ (POYO-1) & $68.48 \pm 7.49$ & $+6.82$ & 7/8 \\
PD & Bal.\ Acc. & $68.12 \pm 7.11$ (BrainOmni) & $67.45 \pm 6.75$ & $+0.68$ & 5/8 \\
\bottomrule
\end{tabular}%
}
\end{table*}

\section{Full Experimental Results}
\label{app:full-results}

\begin{table*}[!t]
\centering
\caption{Full results on the AD EEG benchmark (mean $\pm$ std over 8 seeds). Top section: baselines. Bottom section: \projectname{} with each model across three Controller stages (Initial Run, Retry~1, Retry~2). ``Vanilla'' denotes \projectname{} without pre-trained models. The primary metric is Macro F1. \textit{Eff.}\ is end-to-end wall-clock time and already includes the Codex baseline run that sets the early-stopping threshold.}\label{tab:ad_full}
\small
\renewcommand{\arraystretch}{1.15}
\resizebox{\textwidth}{!}{%
\begin{tabular}{l cccccccc}
\toprule
\textbf{Method} & \textbf{Macro F1 (\%)} & \textbf{Bal.\ Acc.\ (\%)} & \textbf{Acc.\ (\%)} & \textbf{AUROC (\%)} & \textbf{AD Rec.\ (\%)} & \textbf{Healthy Rec.\ (\%)} & \textbf{FTD Rec.\ (\%)} & \textbf{Eff.\ (s)} \\
\midrule
\multicolumn{9}{l}{\textit{Baselines}} \\
AutoML-Agent      & $5.55 \pm 15.71$   & $5.90 \pm 16.69$   & $6.95 \pm 19.64$   & $9.38 \pm 26.52$   & $9.38 \pm 26.52$   & $0.00 \pm 0.00$    & $8.33 \pm 23.57$   & $255$  \\
LAMBDA            & $43.57 \pm 38.28$  & $44.06 \pm 38.57$  & $45.37 \pm 39.33$  & $54.99 \pm 45.83$  & $33.61 \pm 36.35$  & $50.00 \pm 53.45$  & $11.07 \pm 12.16$  & $100$  \\
AutoGen           & $26.75 \pm 11.70$  & $27.25 \pm 15.03$  & $32.75 \pm 10.59$  & $44.25 \pm 19.95$  & $52.00 \pm 25.14$  & $20.38 \pm 19.70$  & $5.00 \pm 14.14$   & $103$  \\
Claude-Sonnet-4.5 & $48.70 \pm 6.58$   & $48.63 \pm 6.97$   & $51.00 \pm 7.18$   & $68.47 \pm 3.27$   & $63.15 \pm 10.48$  & $61.91 \pm 12.36$  & $30.86 \pm 13.47$  & $304$  \\
Codex             & $51.01 \pm 2.52$   & $51.24 \pm 2.75$   & $53.98 \pm 2.98$   & $69.81 \pm 1.15$   & $64.58 \pm 4.64$   & $62.50 \pm 3.88$   & $26.63 \pm 3.63$   & $68$   \\
\midrule
\multicolumn{9}{l}{\textit{\projectname{} (Vanilla)}} \\
Initial Run   & $49.80 \pm 3.29$   & $52.21 \pm 2.54$   & $55.46 \pm 2.32$   & $69.15 \pm 3.69$   & $67.66 \pm 3.58$   & $71.51 \pm 4.40$   & $17.71 \pm 6.72$   & $440$  \\
Retry 1       & $51.15 \pm 5.74$   & $53.20 \pm 4.24$   & $55.21 \pm 4.70$   & $68.09 \pm 6.59$   & $66.12 \pm 6.59$   & $60.72 \pm 19.09$  & $32.29 \pm 20.14$  & $385$  \\
Retry 2       & $47.47 \pm 9.53$   & $49.64 \pm 9.11$   & $51.77 \pm 8.59$   & $66.65 \pm 6.87$   & $65.34 \pm 8.79$   & $59.95 \pm 18.25$  & $30.50 \pm 22.15$  & $252$  \\
\midrule
\multicolumn{9}{l}{\textit{\projectname{} (LaBraM)}} \\
Initial Run   & $28.39 \pm 3.89$   & $29.54 \pm 4.00$   & $31.11 \pm 4.18$   & $44.95 \pm 4.65$   & $32.99 \pm 14.72$  & $25.86 \pm 12.23$  & $17.92 \pm 9.13$   & $486$  \\
Retry 1       & $31.47 \pm 6.98$   & $32.50 \pm 6.86$   & $33.88 \pm 7.52$   & $47.10 \pm 8.34$   & $38.98 \pm 8.87$   & $36.11 \pm 11.97$  & $23.02 \pm 5.28$   & $128$  \\
Retry 2       & $26.21 \pm 3.79$   & $27.20 \pm 3.91$   & $28.16 \pm 3.52$   & $41.73 \pm 4.02$   & $37.09 \pm 9.04$   & $32.85 \pm 13.62$  & $20.52 \pm 7.12$   & $344$  \\
\midrule
\multicolumn{9}{l}{\textit{\projectname{} (REVE)}} \\
Initial Run   & $52.95 \pm 4.49$   & $54.16 \pm 4.89$   & $56.11 \pm 5.20$   & $71.85 \pm 3.19$   & $60.79 \pm 7.01$   & $68.61 \pm 5.90$   & $32.99 \pm 4.45$   & $236$  \\
Retry 1       & $51.05 \pm 6.23$   & $52.40 \pm 6.20$   & $54.08 \pm 6.73$   & $69.26 \pm 4.49$   & $58.06 \pm 8.22$   & $66.75 \pm 6.37$   & $32.38 \pm 5.59$   & $296$  \\
Retry 2       & $50.04 \pm 5.24$   & $51.21 \pm 5.40$   & $52.86 \pm 5.45$   & $67.81 \pm 3.21$   & $57.00 \pm 4.92$   & $65.23 \pm 8.26$   & $31.50 \pm 7.23$   & $190$  \\
\midrule
\multicolumn{9}{l}{\textit{\projectname{} (CBraMod)}} \\
Initial Run   & $34.59 \pm 9.74$   & $36.23 \pm 9.31$   & $36.10 \pm 9.92$   & $50.86 \pm 10.24$  & $39.62 \pm 12.80$  & $36.58 \pm 14.35$  & $30.61 \pm 7.85$   & $469$  \\
Retry 1       & $35.17 \pm 10.32$  & $37.28 \pm 9.14$   & $38.08 \pm 10.00$  & $53.89 \pm 8.40$   & $42.50 \pm 13.16$  & $43.06 \pm 17.73$  & $25.69 \pm 9.78$   & $262$  \\
Retry 2       & $38.06 \pm 6.91$   & $39.86 \pm 5.94$   & $40.97 \pm 6.37$   & $56.59 \pm 5.72$   & $42.96 \pm 9.71$   & $49.32 \pm 13.40$  & $26.35 \pm 9.71$   & $281$  \\
\midrule
\multicolumn{9}{l}{\textit{\projectname{} (EEGPT)}} \\
Initial Run   & $48.83 \pm 3.13$   & $49.91 \pm 2.89$   & $50.99 \pm 2.24$   & $67.29 \pm 2.41$   & $54.10 \pm 2.75$   & $58.53 \pm 6.39$   & $36.58 \pm 12.68$  & $296$  \\
Retry 1       & $48.34 \pm 4.17$   & $49.56 \pm 4.56$   & $50.69 \pm 3.92$   & $66.70 \pm 3.44$   & $52.96 \pm 2.76$   & $59.53 \pm 6.24$   & $36.03 \pm 12.11$  & $175$  \\
Retry 2       & $46.46 \pm 11.00$  & $48.31 \pm 9.30$   & $48.96 \pm 9.91$   & $65.58 \pm 8.83$   & $50.38 \pm 10.70$  & $53.96 \pm 18.65$  & $40.50 \pm 7.73$   & $170$  \\
\midrule
\multicolumn{9}{l}{\textit{\projectname{} (EEGMamba)}} \\
Initial Run   & $30.11 \pm 4.42$   & $31.39 \pm 4.67$   & $32.42 \pm 4.33$   & $50.09 \pm 3.23$   & $39.61 \pm 9.23$   & $29.69 \pm 3.21$   & $24.60 \pm 8.56$   & $475$  \\
Retry 1       & $34.10 \pm 5.41$   & $35.14 \pm 5.58$   & $35.89 \pm 5.42$   & $52.12 \pm 5.33$   & $41.81 \pm 9.77$   & $34.12 \pm 7.56$   & $28.79 \pm 7.32$   & $527$  \\
Retry 2       & $35.74 \pm 6.53$   & $36.27 \pm 5.97$   & $37.39 \pm 5.81$   & $54.17 \pm 5.27$   & $43.54 \pm 9.73$   & $37.12 \pm 6.11$   & $28.17 \pm 8.91$   & $569$  \\
\midrule
\multicolumn{9}{l}{\textit{\projectname{} (BrainOmni)}} \\
Initial Run   & $20.76 \pm 28.79$  & $21.09 \pm 29.22$  & $21.60 \pm 29.92$  & $28.34 \pm 39.11$  & $22.23 \pm 30.72$  & $25.44 \pm 35.41$  & $15.76 \pm 22.29$  & $864$  \\
Retry 1       & $28.82 \pm 30.84$  & $29.24 \pm 31.27$  & $29.97 \pm 32.05$  & $38.33 \pm 40.98$  & $30.90 \pm 33.11$  & $35.79 \pm 38.32$  & $21.20 \pm 23.41$  & $642$  \\
Retry 2       & $27.28 \pm 29.44$  & $27.96 \pm 30.07$  & $28.55 \pm 30.74$  & $36.71 \pm 39.51$  & $29.45 \pm 31.98$  & $33.40 \pm 36.25$  & $21.06 \pm 23.23$  & $597$  \\
\midrule
\multicolumn{9}{l}{\textit{\projectname{} (Auto Mode)}} \\
Best model    & $\mathbf{57.16 \pm 5.71}$   & $\mathbf{58.52 \pm 5.30}$   & $\mathbf{60.57 \pm 4.32}$   & $\mathbf{75.09 \pm 4.83}$   & $\mathbf{66.18 \pm 3.62}$   & $\mathbf{72.97 \pm 4.74}$   & $\mathbf{36.25 \pm 14.86}$  & $425$  \\
\bottomrule
\end{tabular}%
}
\end{table*}

\begin{samepage}
Tables~\ref{tab:ad_full} and~\ref{tab:pd_full} present the complete per-model, per-stage results
on the AD and PD benchmarks, including all baseline comparisons and the iterative
Controller optimization trajectory for each pre-trained model. WM results are
reported in Table~\ref{tab:main_results} for the baseline comparison and in the WM row of
Table~\ref{tab:heldout} for the held-out evaluation.
\end{samepage}

\begin{table*}[!ht]
\centering
\caption{Full results on the PD EEG benchmark (mean $\pm$ std over 8 seeds). Top section: baselines. Bottom section: \projectname{} with each model across three Controller stages (Initial Run, Retry~1, Retry~2). ``Vanilla'' denotes \projectname{} without pre-trained models. The primary metric is Balanced Accuracy. Note that Claude-Sonnet-4.5 achieves the highest PD Recall but at the cost of substantially lower Control Recall (46.28\%), indicating a strong prediction bias toward the majority class. \textit{Eff.}\ is end-to-end wall-clock time and already includes the Codex baseline run that sets the early-stopping threshold.}\label{tab:pd_full}
\small
\renewcommand{\arraystretch}{1.15}
\resizebox{\textwidth}{!}{%
\begin{tabular}{l ccccccc}
\toprule
\textbf{Method} & \textbf{Bal.\ Acc.\ (\%)} & \textbf{Acc.\ (\%)} & \textbf{Macro F1 (\%)} & \textbf{AUROC (\%)} & \textbf{Control Rec.\ (\%)} & \textbf{PD Rec.\ (\%)} & \textbf{Eff.\ (s)} \\
\midrule
\multicolumn{8}{l}{\textit{Baselines}} \\
AutoML-Agent      & $17.39 \pm 32.29$  & $18.33 \pm 33.95$  & $9.15 \pm 25.88$   & $19.90 \pm 37.01$  & $10.71 \pm 30.30$  & $7.81 \pm 22.10$   & $293$  \\
LAMBDA            & $0.00 \pm 0.00$    & $0.00 \pm 0.00$    & $0.00 \pm 0.00$    & $0.00 \pm 0.00$    & $0.00 \pm 0.00$    & $0.00 \pm 0.00$    & $284$  \\
AutoGen           & $6.25 \pm 17.68$   & $8.38 \pm 23.69$   & $5.00 \pm 14.14$   & $6.25 \pm 17.68$   & $0.00 \pm 0.00$    & $12.50 \pm 35.36$  & $308$  \\
Claude-Sonnet-4.5 & $65.12 \pm 3.27$   & $70.99 \pm 1.54$   & $65.50 \pm 3.43$   & $74.07 \pm 3.52$   & $46.28 \pm 12.37$  & $\mathbf{84.50 \pm 7.53}$  & $516$  \\
Codex             & $65.52 \pm 7.98$   & $64.95 \pm 12.87$  & $62.08 \pm 12.98$  & $71.07 \pm 8.23$   & $64.57 \pm 10.11$  & $61.00 \pm 22.34$  & $125$  \\
\midrule
\multicolumn{8}{l}{\textit{\projectname{} (Vanilla)}} \\
Initial Run   & $64.52 \pm 1.25$   & $70.49 \pm 1.18$   & $64.40 \pm 0.66$   & $74.41 \pm 0.93$   & $47.34 \pm 7.57$   & $81.82 \pm 5.33$   & $113$  \\
Retry 1       & $59.09 \pm 8.50$   & $61.65 \pm 12.15$  & $53.99 \pm 13.74$  & $64.74 \pm 7.77$   & $51.57 \pm 30.95$  & $66.58 \pm 28.92$  & $156$  \\
Retry 2       & $60.48 \pm 7.08$   & $61.59 \pm 12.23$  & $56.08 \pm 13.43$  & $66.14 \pm 7.78$   & $57.16 \pm 23.88$  & $63.75 \pm 27.14$  & $158$  \\
\midrule
\multicolumn{8}{l}{\textit{\projectname{} (LaBraM)}} \\
Initial Run   & $59.48 \pm 0.56$   & $61.08 \pm 0.65$   & $58.35 \pm 0.54$   & $60.20 \pm 1.32$   & $55.10 \pm 0.00$   & $64.08 \pm 0.94$   & $70$   \\
Retry 1       & $56.49 \pm 3.81$   & $57.93 \pm 3.99$   & $55.10 \pm 3.86$   & $58.18 \pm 4.17$   & $52.21 \pm 3.46$   & $60.73 \pm 4.34$   & $76$   \\
Retry 2       & $60.25 \pm 1.47$   & $61.58 \pm 1.57$   & $58.75 \pm 1.33$   & $60.70 \pm 1.13$   & $56.72 \pm 2.57$   & $64.02 \pm 2.28$   & $95$   \\
\midrule
\multicolumn{8}{l}{\textit{\projectname{} (REVE)}} \\
Initial Run   & $62.83 \pm 0.16$   & $68.48 \pm 0.01$   & $63.06 \pm 0.10$   & $63.11 \pm 0.61$   & $46.56 \pm 0.53$   & $79.19 \pm 0.26$   & $78$   \\
Retry 1       & $56.47 \pm 6.13$   & $61.08 \pm 8.02$   & $56.27 \pm 6.53$   & $58.92 \pm 7.72$   & $43.07 \pm 1.94$   & $69.85 \pm 11.56$  & $90$   \\
Retry 2       & $58.92 \pm 4.40$   & $63.91 \pm 5.75$   & $58.73 \pm 4.69$   & $60.98 \pm 5.29$   & $44.39 \pm 2.56$   & $73.44 \pm 8.45$   & $117$  \\
\midrule
\multicolumn{8}{l}{\textit{\projectname{} (CBraMod)}} \\
Initial Run   & $58.72 \pm 0.42$   & $60.70 \pm 0.51$   & $57.37 \pm 0.51$   & $60.73 \pm 0.66$   & $53.23 \pm 0.31$   & $64.42 \pm 0.83$   & $85$   \\
Retry 1       & $59.58 \pm 3.95$   & $61.64 \pm 5.26$   & $58.11 \pm 4.60$   & $62.64 \pm 4.16$   & $53.36 \pm 3.95$   & $65.67 \pm 8.29$   & $87$   \\
Retry 2       & $60.55 \pm 1.57$   & $62.95 \pm 1.46$   & $59.24 \pm 1.41$   & $63.86 \pm 3.26$   & $53.23 \pm 3.37$   & $67.64 \pm 2.38$   & $97$   \\
\midrule
\multicolumn{8}{l}{\textit{\projectname{} (EEGPT)}} \\
Initial Run   & $65.72 \pm 1.24$   & $68.38 \pm 1.45$   & $65.16 \pm 1.22$   & $69.22 \pm 1.41$   & $57.69 \pm 0.60$   & $73.69 \pm 1.81$   & $91$   \\
Retry 1       & $58.42 \pm 10.01$  & $59.90 \pm 10.47$  & $57.21 \pm 10.18$  & $61.42 \pm 13.44$  & $53.91 \pm 9.66$   & $62.88 \pm 11.73$  & $106$  \\
Retry 2       & $61.08 \pm 8.28$   & $62.57 \pm 8.29$   & $59.86 \pm 8.23$   & $64.91 \pm 10.97$  & $56.67 \pm 9.28$   & $65.48 \pm 8.76$   & $126$  \\
\midrule
\multicolumn{8}{l}{\textit{\projectname{} (EEGMamba)}} \\
Initial Run   & $56.14 \pm 0.39$   & $57.92 \pm 0.38$   & $54.81 \pm 0.41$   & $57.79 \pm 0.60$   & $51.19 \pm 0.31$   & $61.27 \pm 0.52$   & $96$   \\
Retry 1       & $48.31 \pm 4.86$   & $50.35 \pm 5.12$   & $46.96 \pm 5.15$   & $49.80 \pm 6.61$   & $42.22 \pm 5.46$   & $54.31 \pm 5.91$   & $77$   \\
Retry 2       & $52.11 \pm 2.65$   & $56.11 \pm 5.26$   & $50.39 \pm 3.00$   & $52.84 \pm 3.59$   & $40.31 \pm 13.97$  & $63.81 \pm 13.76$  & $104$  \\
\midrule
\multicolumn{8}{l}{\textit{\projectname{} (BrainOmni)}} \\
Initial Run   & $67.29 \pm 2.44$   & $71.39 \pm 1.88$   & $67.21 \pm 2.46$   & $73.50 \pm 1.77$   & $55.36 \pm 4.07$   & $79.21 \pm 1.30$   & $114$  \\
Retry 1       & $49.99 \pm 21.81$  & $51.12 \pm 22.86$  & $48.78 \pm 21.68$  & $52.36 \pm 23.91$  & $46.64 \pm 19.40$  & $53.29 \pm 25.15$  & $74$   \\
Retry 2       & $56.28 \pm 7.41$   & $57.83 \pm 8.71$   & $54.99 \pm 7.94$   & $61.15 \pm 11.79$  & $51.74 \pm 5.36$   & $60.79 \pm 11.54$  & $105$  \\
\midrule
\multicolumn{8}{l}{\textit{\projectname{} (Auto Mode)}} \\
Best model    & $\mathbf{71.32 \pm 2.97}$   & $\mathbf{71.91 \pm 1.87}$   & $\mathbf{69.22 \pm 2.47}$   & $\mathbf{78.15 \pm 1.21}$   & $\mathbf{69.80 \pm 11.88}$  & $72.95 \pm 7.27$  & $975$ \\
\bottomrule
\end{tabular}%
}
\end{table*}

\section{Timeouts and the BrainOmni Re-run}
\label{app:timeouts}

Table~\ref{tab:ad_full} reports BrainOmni's initial round on AD as $20.76 \pm 28.79$
Macro F1. A standard deviation larger than the mean is the signature of a budget failure rather
than a modeling one: several seeds produced no score at all because the script exceeded the
300\,s execution limit, and those seeds enter the average as zeros. BrainOmni has the largest
embedding dimension of the nine models we integrate (4096; Table~\ref{tab:models}), so it is the
model this uniform budget binds hardest.

We re-ran BrainOmni on AD with the model fixed in advance and a more generous
allocation, over the same eight seeds. Table~\ref{tab:brainomni_rerun} gives the result. Every
metric roughly triples, but the informative change is the dispersion: each standard deviation falls
by a factor of eight or more, which is what one expects when zeros stop entering the average.
The re-runs completed in $318$--$680$\,s per seed and every seed produced a
score, which is what the collapse in dispersion reflects. We omit the three per-class recall columns of
Table~\ref{tab:ad_full}: the generated scripts record them under inconsistent keys across seeds,
and we could not establish a single class ordering we trust for all eight.

\begin{table}[H]
\centering
\small
\caption{BrainOmni on AD under the standard budget, as reported in
Table~\ref{tab:ad_full}, and re-run with a more generous allocation over the same eight seeds.
Every standard deviation falls by a factor of eight or more.}
\label{tab:brainomni_rerun}
\begin{tabular}{lcc}
\toprule
\textbf{Metric} & \textbf{Standard budget} & \textbf{Re-run} \\
\midrule
Macro F1 (\%)   & $20.76 \pm 28.79$ & $\mathbf{62.32 \pm 3.27}$ \\
Bal.\ Acc.\ (\%) & $21.09 \pm 29.22$ & $\mathbf{62.85 \pm 3.49}$ \\
Acc.\ (\%)       & $21.60 \pm 29.92$ & $\mathbf{64.17 \pm 3.31}$ \\
AUROC (\%)      & $28.34 \pm 39.11$ & $\mathbf{81.43 \pm 1.90}$ \\
\bottomrule
\end{tabular}
\end{table}

This also resolves the cross-dataset discrepancy for this model. Ranked by the initial
round under each dataset's own primary metric, BrainOmni places last of seven on AD but first of
seven on PD, which is not what one expects from the same pre-trained model on two resting-state EEG
corpora. With the timeout lifted it places first on AD as well, and the two datasets agree. The
discrepancy was a property of the compute budget, not of the model.

We report the original figure in Table~\ref{tab:ad_full} rather than substituting the
re-run, because the table describes what the system produces under the budget every model is given.
The re-run is reported here so that the number is not read as a statement about BrainOmni itself.

\section{Web Interface}
\label{app:web-ui}

\begin{figure*}[!t]
\centering
\includegraphics[width=\textwidth, height=0.4\textheight, keepaspectratio]{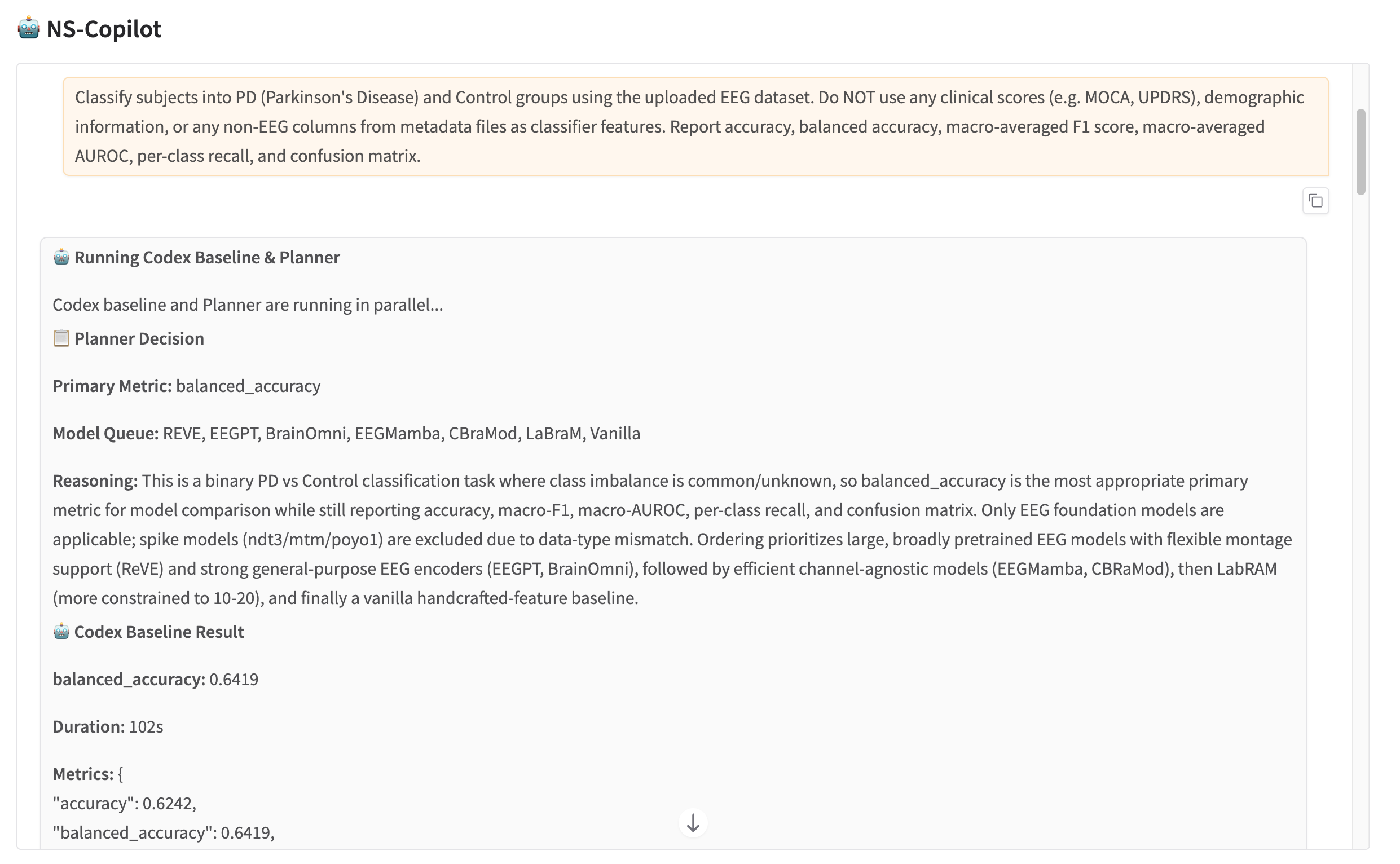}
\caption{User input with task description, parallel performance threshold computation and Planner execution, and Planner's model queue with reasoning.}
\label{fig:web-ui-1}
\end{figure*}

\begin{figure*}[!t]
\centering
\includegraphics[width=\textwidth, height=0.4\textheight, keepaspectratio]{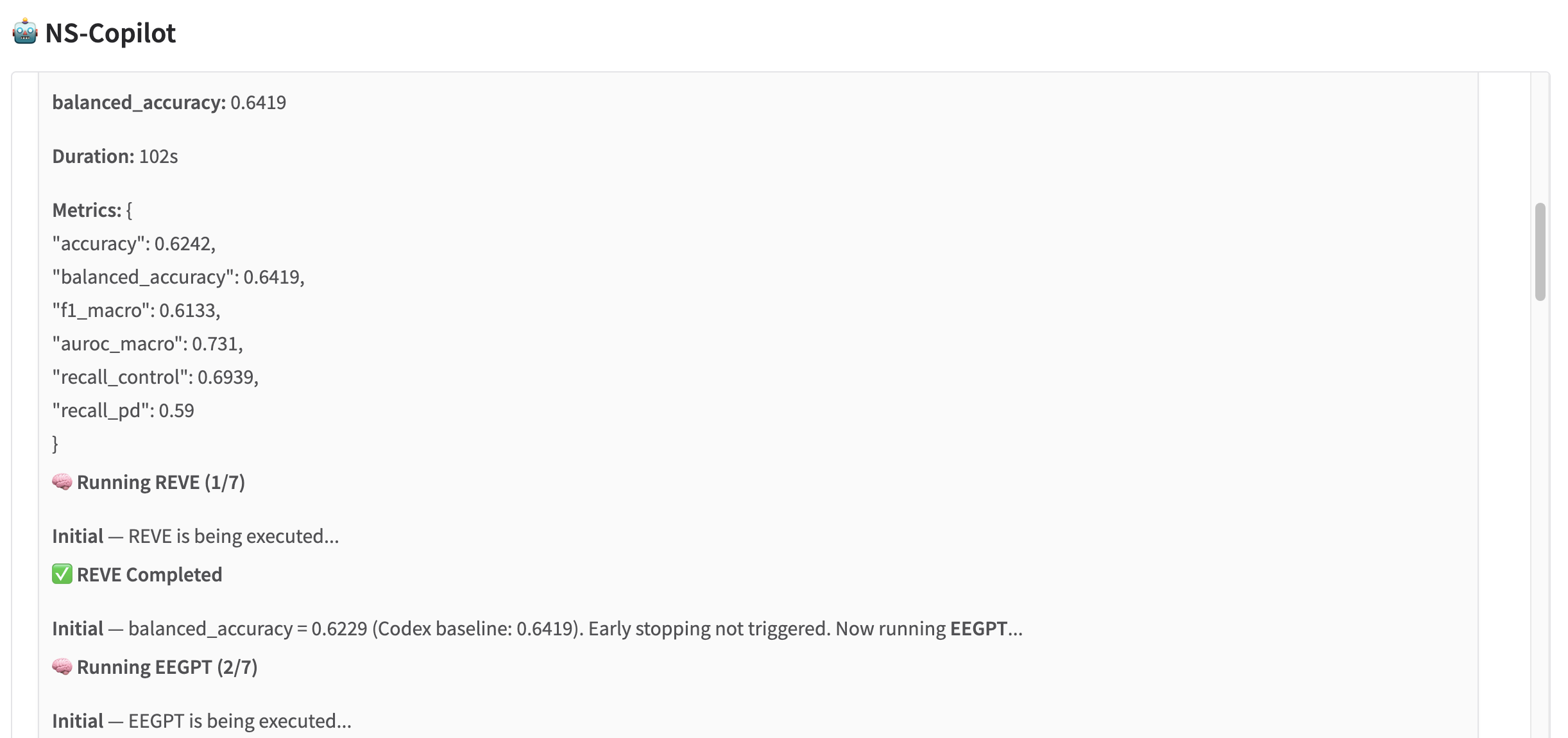}
\caption{Performance threshold metrics and initial model search (REVE followed by EEGPT).}
\label{fig:web-ui-2}
\end{figure*}

\begin{figure*}[!t]
\centering
\includegraphics[width=\textwidth, height=0.4\textheight, keepaspectratio]{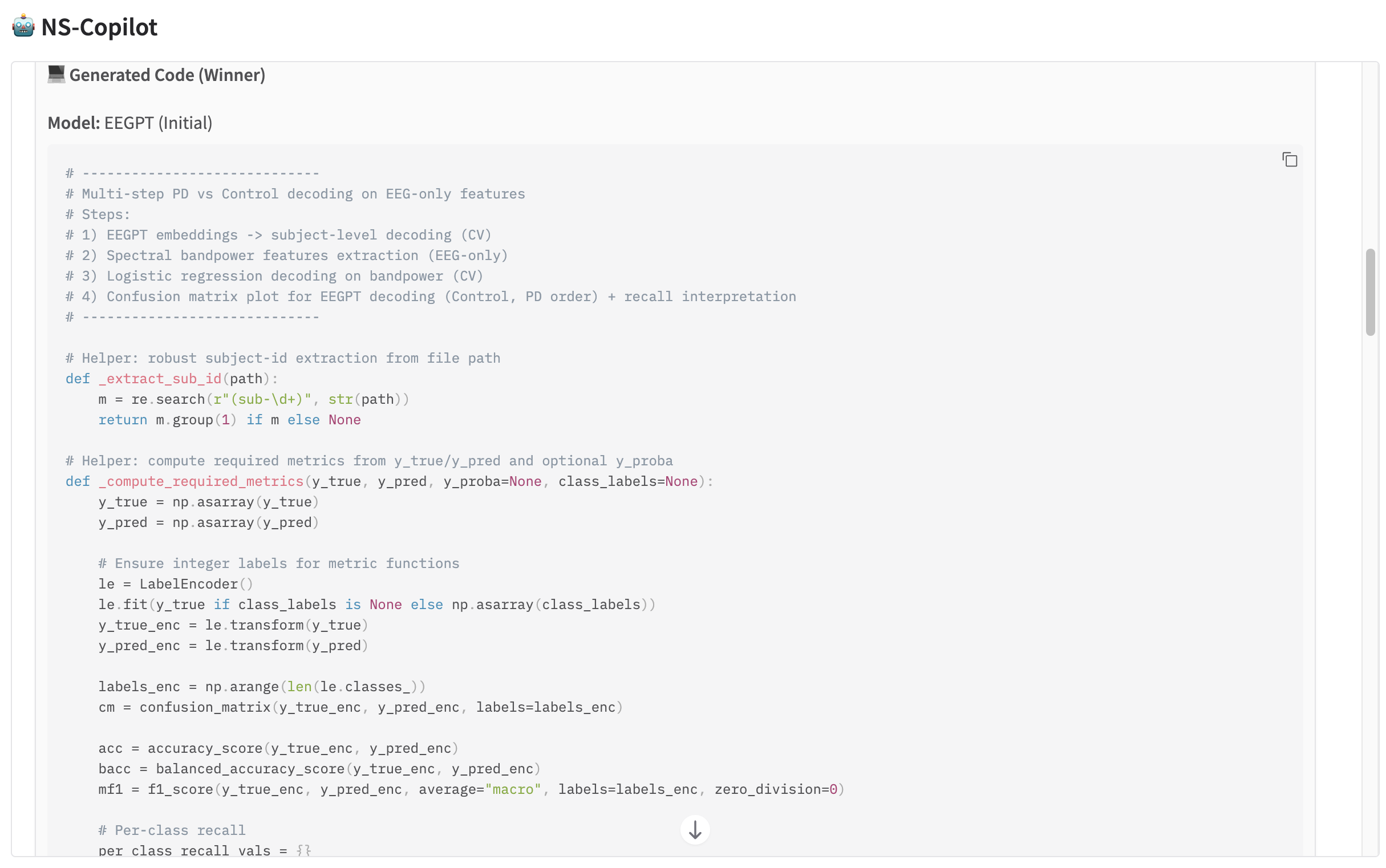}
\caption{Generated Python code for the winning model (EEGPT).}
\label{fig:web-ui-3}
\end{figure*}

\begin{figure*}[!t]
\centering
\includegraphics[width=\textwidth, height=0.4\textheight, keepaspectratio]{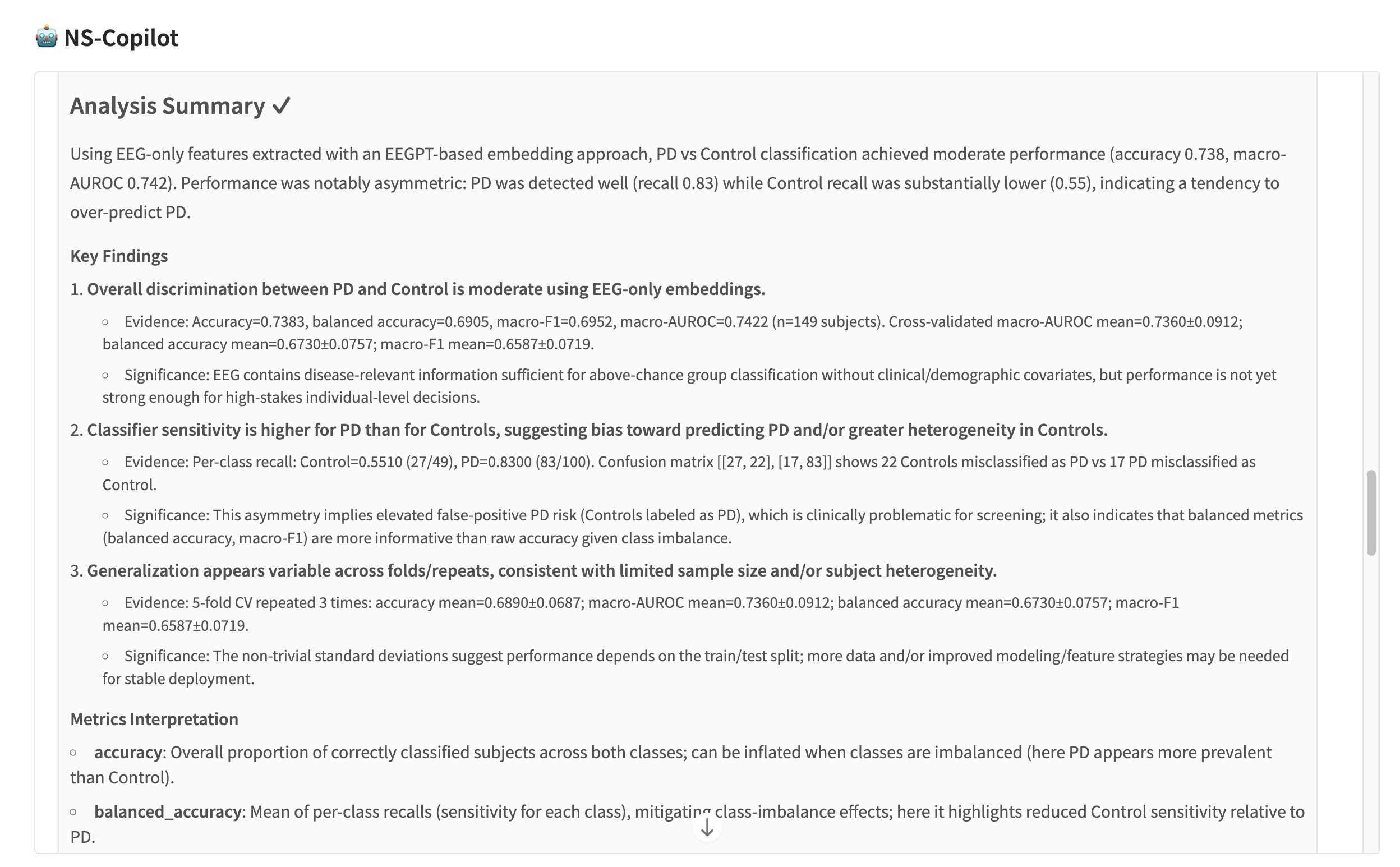}
\caption{Interpreter analysis summary with key findings and metric interpretation.}
\label{fig:web-ui-4}
\end{figure*}

\begin{figure*}[!t]
\centering
\includegraphics[width=\textwidth, height=0.4\textheight, keepaspectratio]{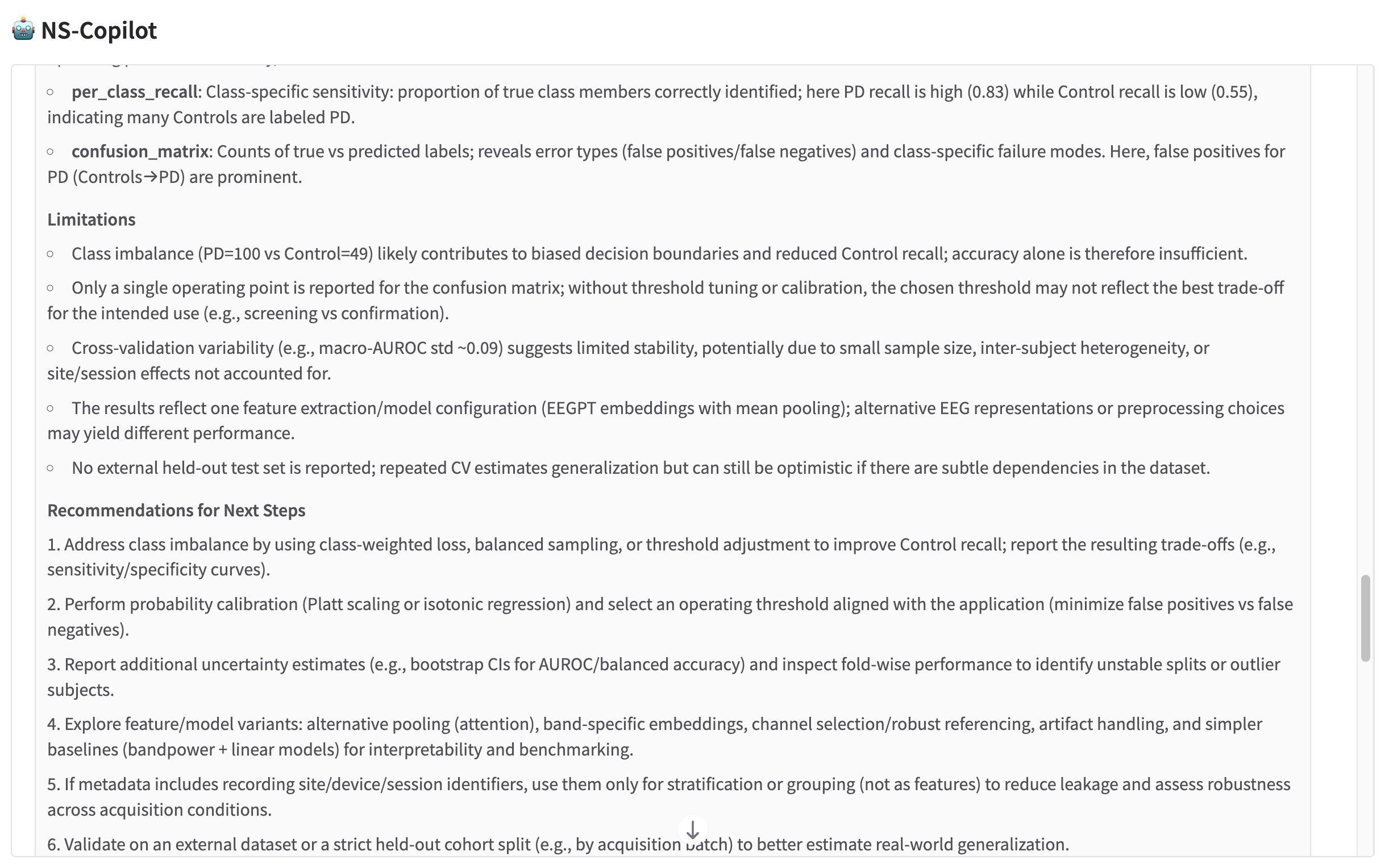}
\caption{Interpreter limitations and recommendations for next steps.}
\label{fig:web-ui-5}
\end{figure*}

\begin{figure*}[!t]
\centering
\includegraphics[width=\textwidth, height=0.4\textheight, keepaspectratio]{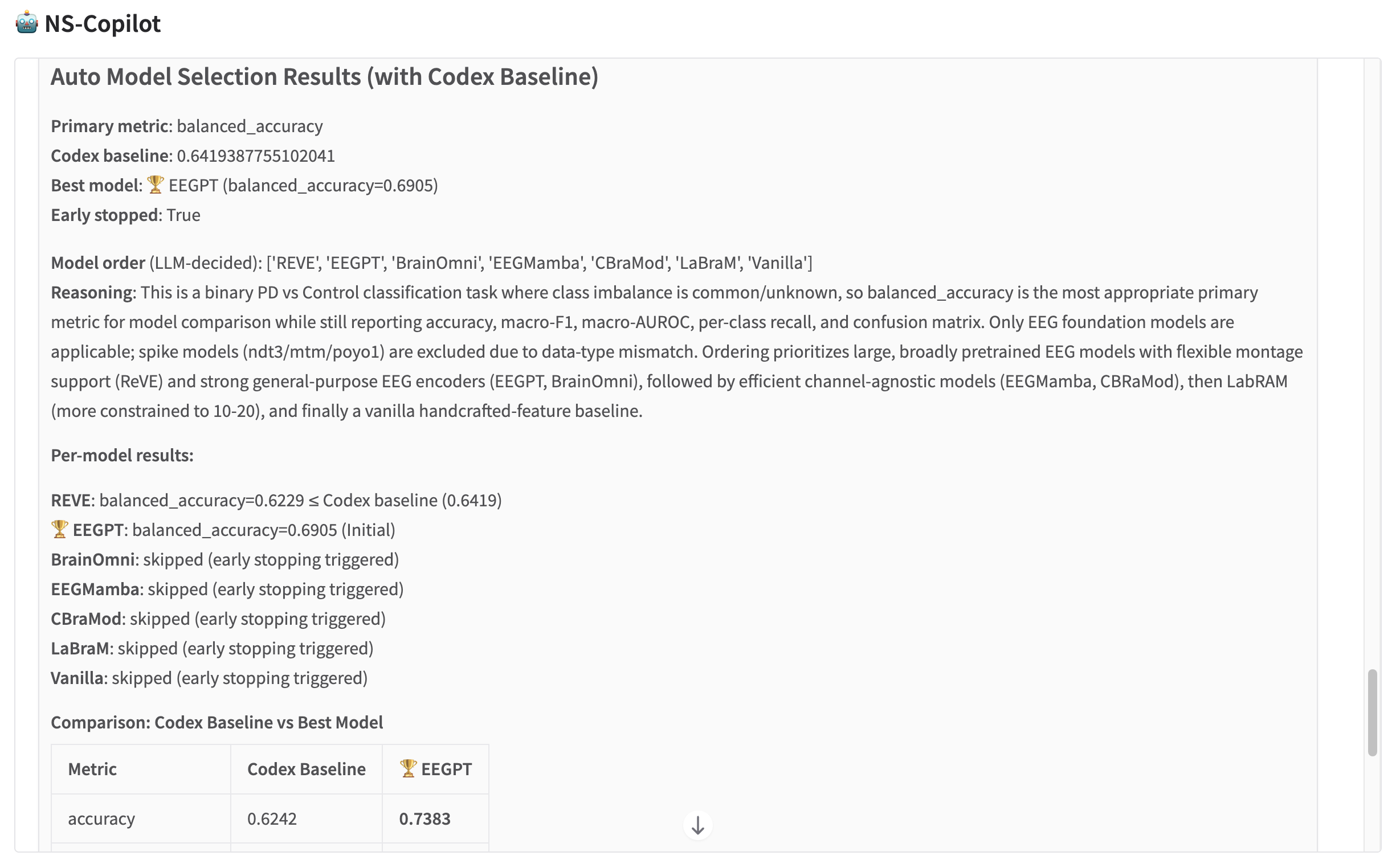}
\caption{Auto Model Selection results with per-model execution outcomes, and a comparison table between the winner model and the performance threshold.}
\label{fig:web-ui-6}
\end{figure*}

\begin{figure*}[!t]
\centering
\includegraphics[width=\textwidth, height=0.45\textheight, keepaspectratio]{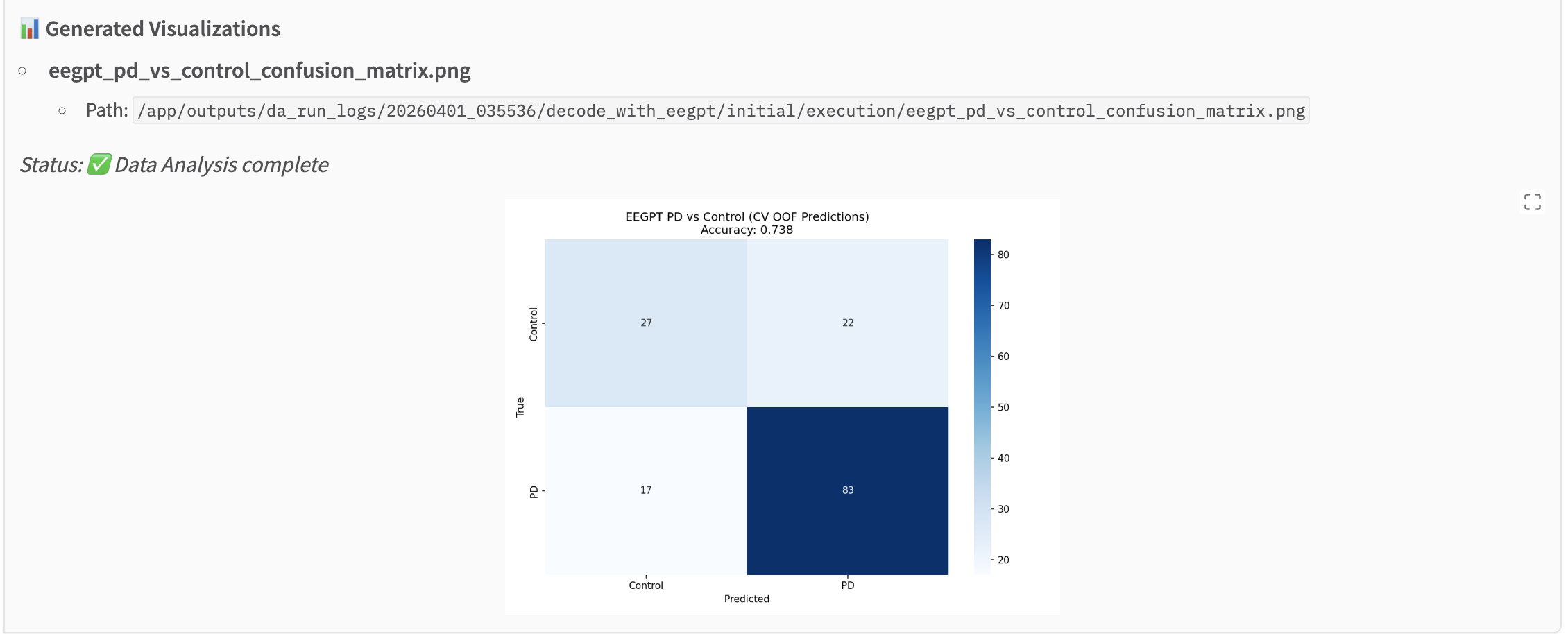}
\caption{\projectname{} web interface: generated confusion matrix visualization and task completion status.}
\label{fig:web-ui-7}
\end{figure*}

Figures~\ref{fig:web-ui-1}--\ref{fig:web-ui-7} illustrate the full \projectname{} web interface workflow on the PD EEG classification task.

\end{document}